\documentclass[letterpaper, 10 pt, conference]{ieeeconf}  
\usepackage{cite}
\usepackage{graphicx}
\usepackage{tcolorbox}
\usepackage{xcolor}
\usepackage{hyperref}
\usepackage[]{footmisc}
\usepackage{adjustbox}
\usepackage{booktabs,graphicx,makecell,siunitx,eqparbox}
\usepackage{tabularx}
\usepackage{footnote}
\usepackage{balance}
\usepackage{svg}
\usepackage{caption}
\usepackage{subcaption}
\usepackage{amssymb}
\usepackage{amsmath}
\usepackage{algorithm}
\usepackage{algpseudocode}
\makesavenoteenv{tabular}
\usepackage{cuted}
\usepackage{wrapfig}
\usepackage{caption}
\usepackage{graphicx} 
\usepackage{booktabs} 
\usepackage{pifont}   
\usepackage{tabularx}
\usepackage{array}
\usepackage{pifont}
\usepackage{xcolor}
\IEEEoverridecommandlockouts                              

\newcommand{\equalcontrib}{\textsuperscript{*}}
\newcommand{\corresponding}{\textsuperscript{\textdagger}}

\title{\LARGE\bfseries
VidAct: Learning Manipulation from In-the-Wild Videos \\
with Object-Centric 3D Awareness
}

\author{
\mdseries
Hang Li\textsuperscript{1}\equalcontrib
\quad
Mingxin Zhang\textsuperscript{1}\equalcontrib
\quad
Zihan Wu\textsuperscript{1}\equalcontrib
\quad
Yang Tian\textsuperscript{2}
\quad
Dong Chen\textsuperscript{3}
\quad
Fengyi Shen\textsuperscript{1,3}
\quad
Yuan Meng\textsuperscript{1}
\\
Xiangtong Yao\textsuperscript{1}
\quad
Heng Zhang\textsuperscript{4}
\quad
Ziyuan Liu\textsuperscript{3}\corresponding
\quad
Zhenshan Bing\textsuperscript{5,1}\corresponding
\quad
Alois Knoll\textsuperscript{1}
\\[5pt]
\textsuperscript{1}Technical University of Munich
\quad
\textsuperscript{2}Peking University
\quad
\textsuperscript{3}Huawei Heisenberg Research Center
\\
\textsuperscript{4}Huawei CloudRobo Lab
\quad
\textsuperscript{5}Nanjing University
}

\begin{document}

\maketitle

\begingroup
\renewcommand{\thefootnote}{}
\footnotetext{%
\textsuperscript{*}Equal contribution, order randomized.
\qquad
\textsuperscript{\textdagger}Corresponding authors.%
}
\endgroup

\thispagestyle{empty}
\pagestyle{empty}

\begin{strip}
    \centering
    \includegraphics[width=\textwidth]{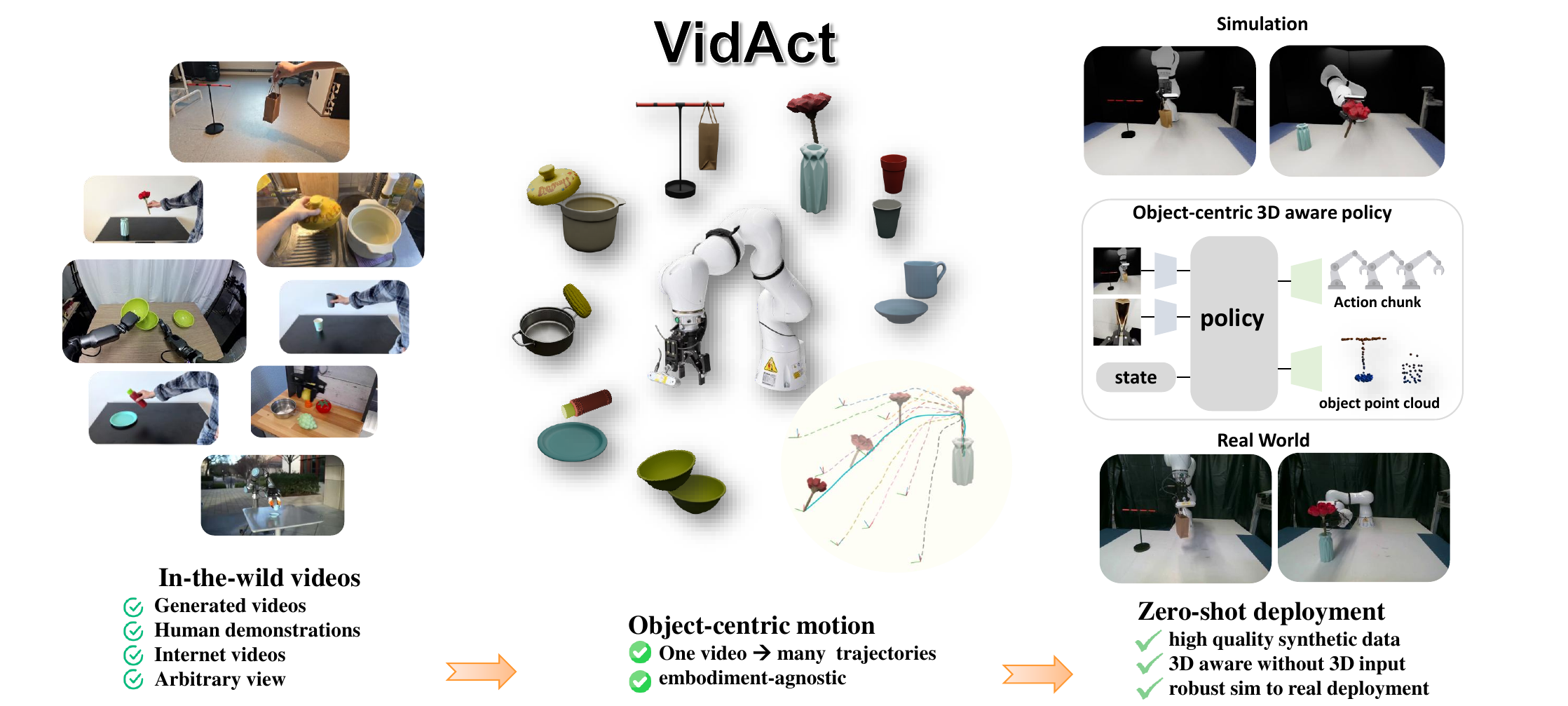}
    \vspace{-1.5em}
    \captionof{figure}{We introduce VidAct, a video-to-robot framework that learns object-centric 3D-aware policies from a single in-the-wild monocular video per task by extracting embodiment- and viewpoint-invariant object motion, augmenting reconstructed trajectories with residual transfer, and generating high-fidelity synthetic data to train policies with privileged current-frame object point-cloud supervision for robust zero-shot deployment using only RGB inputs.}
    \label{fig:overview}
\end{strip}

\begin{abstract}
Video demonstrations offer a scalable alternative to costly robot data for learning manipulation, yet existing reconstruction-based approaches often rely on constrained camera viewpoints or human-to-robot retargeting, while the reconstructed trajectories are difficult to adapt to new objects configurations without distorting the trajectory shape. Another key limitation is that the resulting policies often lack precise object-level 3D geometry awareness, limiting object grounding and object shape awareness critical for precise manipulation. To bridge these gaps, we propose VidAct, an efficient video-to-robot framework that learns object-centric, 3D-aware manipulation policies from a single monocular video per task and enables zero-shot real-world deployment. VidAct consists of three key components. First, VidAct reconstructs object meshes and motion from arbitrary demo videos and canonicalizes the motion in the static object frame, avoiding embodiment-specific retargeting and accommodating diverse camera viewpoints. Second, VidAct employ residual trajectory transfer for adapting the reconstructed motion to novel object configurations while preserving its motion shape. Finally, as the key policy-learning component, VidAct predicts simulation-provided privileged complete-object point clouds at each frame as an auxiliary task while retaining RGB-only deployment, providing dense object-centric supervision over both object pose and 3D geometry. Experiments on human, robot, generated, and internet videos demonstrate broad video applicability and zero-shot deployment. Per-frame complete-object 3D supervision improves policy generalization and sim-to-real success, while residual trajectory transfer enables reliable trajectory adaptation with better shape preservation.



\end{abstract}

\section{INTRODUCTION}


Humans can acquire manipulation skills simply by watching videos, regardless of the camera viewpoint or whether the demonstrated actions are performed by a human or a robot. Endowing robots with a similar ability to rapidly learn manipulation skills from videos could facilitate their widespread deployment in the real world. Learning robotic manipulation from videos has become an increasingly active area of research. One line of work leverages large-scale human videos to pretrain generalist policies \cite{kareer2025emergence,EgoScale}, but typically still requires robot data to finetune the model for downstream tasks. Another line converts videos into robot training data via reconstruction, often through Real2Sim2Real pipelines~\cite{xsim,deximit,rigvid}, enabling trained policies to be directly deployed in the real world. Our work falls within the latter line of research and addresses several challenges in this paradigm.

First, reconstructing in-the-wild videos presents several challenges. Existing methods commonly use human-to-robot hand retargeting, requiring clearly visible human hands and coupling the reconstructed motion to a specific robot embodiment. Reconstructed motions are also camera-dependent. In addition, reliable object pose estimation requires metric depth and scaled object meshes that are difficult to obtain from monocular videos. Moreover, a single reconstructed trajectory must be adapted to diverse object configurations while preserving its trajectory shape. Existing methods use interpolation~\cite{real2render2real}, segment replay with motion planning~\cite{demogen,deximit}, or RL~\cite{rialto,xsim}, which can distort the demonstrated motion or require costly task-specific training.

Furthermore, after obtaining robot data from videos, training a standard imitation learning policy may still yield insufficient robustness. Most robot policies take RGB observations as input~\cite{ACT, Diffusion-Policy, pi0.5, ReMem-VLA}, which has limited grounding and awareness of precise object-level 3D geometry. Accurate object grounding and complete 3D shape are critical for fine-grained manipulation, such as insertion and stacking. Prior work has used 3D point clouds as input~\cite{3D-Diffusion-Policy, o3dp}, which are noisy and partially observable in the real world. Some methods require costly manual 2D grounding annotations~\cite{dreamvla, robointer}, while precise 3D grounding and object geometry are difficult to annotate from RGB. Simulation provides accurate, noise-free privileged information, including object poses, depth, and mesh-sampled point clouds. How to effectively leverage such information for policy learning and sim-to-real transfer remains underexplored.

To bridge these gaps, we propose VidAct, which converts in-the-wild videos into high-quality robot data for learning object-centric 3D-aware policies deployable zero-shot in the real world with only RGB inputs. As shown in Figure \ref{fig:overview}, VidAct provides three key advances: (1) A Video to Sim pipeline that extracts transferable object motion from arbitrary-view monocular videos. Instead of retargeting human hand motion, we use the reconstructed object pose sequence as the motion reference, making it independent of both the demonstrator and robot embodiment. To remove camera-viewpoint dependence, we express the object motion in static object frame, yielding a camera-invariant trajectory representation. For stable object pose estimation, we align the scale  of reconstructed mesh to the estimated depth scale. (2) A simple and efficient mechanism---residual trajectory transfer---is employed to adapt motions to novel object poses. Transferring motion residuals between start-to-goal baselines enables efficient adaptation while retaining characteristic motion patterns. (3) An object-centric 3D auxiliary supervision strategy using simulation-provided privileged point clouds only during training. At each frame, an auxiliary head predicts the complete point clouds of task-relevant objects, providing dense supervision over both object pose and 3D geometry. Combined with background randomization, this encourages the visual representation to focus on task-relevant object geometry rather than scene-specific appearance, improving generalization to unseen scenes. Since the privileged 3D information is used only as a training target, the deployed policy remains RGB-only, avoiding noisy real-world 3D inputs and an additional sim-to-real gap.

Our main contributions are summarized as follows:
\begin{enumerate}
    \item We propose VidAct, a broadly applicable video-to-robot framework that is independent of demonstrator embodiment, enabling object-centric, 3D-aware policy learning and zero-shot real-world deployment.

    \item We employ residual tranjectory transfer for lightweight adaptation to novel object poses while preserving motion shape, without costly optimization or RL.

    \item We introduce training-only complete-object 3D supervision from simulation to improve generalization of RGB policy and sim-to-real performance.

    \item Experiments on diverse videos and two robot embodiments validate VidAct's applicability and the benefits of residual trajectory transfer and complete-object 3D supervision.
\end{enumerate}
\section{RELATED WORK}
\label{sec:related works}
\subsection{Learning from Videos}
The central challenge is to convert action-free videos into robot-executable supervision. Existing approaches mainly differ in the transferable signal extracted from videos. One line uses large-scale human videos for scalable but indirect supervision with human-hand motion annotations \cite{ kareer2025emergence,EgoScale,egodex}. However, they do not directly provide robot-executable actions and still require robot data for finetuning. Another line reconstruct human motion, such as hand, wrist, or point trajectories, and retargets it to robots \cite{deximit,dexman}. This offers stronger action supervision, but relies on accurate human tracking and embodiment-specific retargeting. A third line shifts the transferable signal from the demonstrator to the manipulated object \cite{xsim,rigvid}, making manipulation more embodiment-independent by defining it through object state changes rather than body motions. However, existing object-centric methods often assume depth sensing or multi-view object scans, limiting their applicability to arbitrary-view monocular videos. Our method extends object-centric motion representation to broader arbitrary-view monocular videos, converting them into robot-executable supervision without restricting the demonstrator to human hands. 

\subsection{Robot Data Synthesis from Reference Motion}
Expanding a single reconstructed trajectory to broader object spatial poses is a key step toward training policies that generalize across object configurations\cite{demogen,mimicgen}. Geometric methods modify the reference trajectory, either by replaying object-relative trajectory segments under newly sampled object poses \cite{mimicgen,dexmimicgen}, or by interpolating trajectories between new start and goal configurations \cite{real2render2real}. These methods are efficient, but they modify the trajectory at the point or fragment level, which can change the demonstrated approach direction, contact timing, or local motion shape. RL-based methods bypass explicit trajectory augmentation by relearning the behavior in simulation from task-specific objectives \cite{rialto,xsim}. They offer greater flexibility, but require reward design and costly optimization, and the resulting behavior is relearned rather than explicitly preserving the reference trajectory. In contrast, our residual trajectory transfer avoids simple interpolation, costly optimization, and manual temporal segmentation into sub-skills, enabling fast pose-level adaptation while preserving the demonstrated motion profile.

We summarize the differences compared with existing baselines in terms of data generation in table~\ref{tab:method_comparison}.
\vspace{-2mm}

\newcommand{\cmark}{\textcolor{green!60!black}{\ding{51}}}
\newcommand{\xmark}{\textcolor{red!75!black}{\ding{55}}}

\begin{table}[h]
\centering
\scriptsize
\setlength{\tabcolsep}{3.0pt}
\caption{Comparison with existing data generation methods.}
\label{tab:method_comparison}
\resizebox{\columnwidth}{!}{
\begin{tabular}{lcccc}
\toprule
Method &
\shortstack{Single \\ RGB Video} &
\shortstack{Embodiment \\ Agnostic RGB Video} &
\shortstack{Retargeting \\ Free} &
\shortstack{RL \\ Free} \\
\midrule
DemoGen \cite{demogen} & \xmark & \xmark & \cmark & \cmark \\
R2R2R \cite{real2render2real} & \xmark & \xmark & \xmark & \cmark \\
X-Sim \cite{xsim} & \xmark & \xmark & \cmark & \xmark \\
RigVid \cite{rigvid} & \xmark & \xmark & \cmark & \cmark \\
DexImit \cite{deximit} & \cmark & \xmark & \xmark & \cmark \\
\textbf{VidAct (Ours)}
        & \cmark & \cmark & \cmark & \cmark \\
\bottomrule
\end{tabular}}
\vspace{-3mm}
\end{table}

VidAct can support both converting human videos to robot data and augmenting robot data from robot videos.

\subsection{Intermediate Representations for Manipulation}
Intermediate representations are widely used to improve manipulation policy learning by grounding actions on task-relevant objects.
One line of work feeds them directly as policy inputs, including object poses, object proposals or bounding boxes~\cite{viola,vima}, object masks or segmentation-conditioned features~\cite{moo,multitask}, object-centric 3D or point-cloud representations~\cite{groot,o3dp}. These inputs provide strong object-level grounding and improve generalization across poses, categories, and layouts, but require reliable perception modules at deployment, such as pose estimation, detection, segmentation, point-cloud reconstruction, which is difficult and unstable on real robots.
Another line uses intermediate representations only as prediction targets or auxiliary supervision, including bounding boxes~\cite{ecot}, optical flow~\cite{flowvla,lamp}, depth~\cite{qdepthvla}, or mixtures of objectives~\cite{dreamvla,robointer}.
This allows deployment from RGB observations without external perception modules, but still depends on dataset annotations, which can be costly, noisy, and hard to scale.
In contrast, our design naturally provides clean object-centric 3D supervision from simulation while keeping the deployed policy RGB-only, enabling object-level 3D awareness without noisy real-world 3D inputs or additional sim-to-real input mismatch.
\section{Approach}
\label{sec:Approach}
\begin{figure*}[t]
    \centering
    \includegraphics[width=\textwidth]{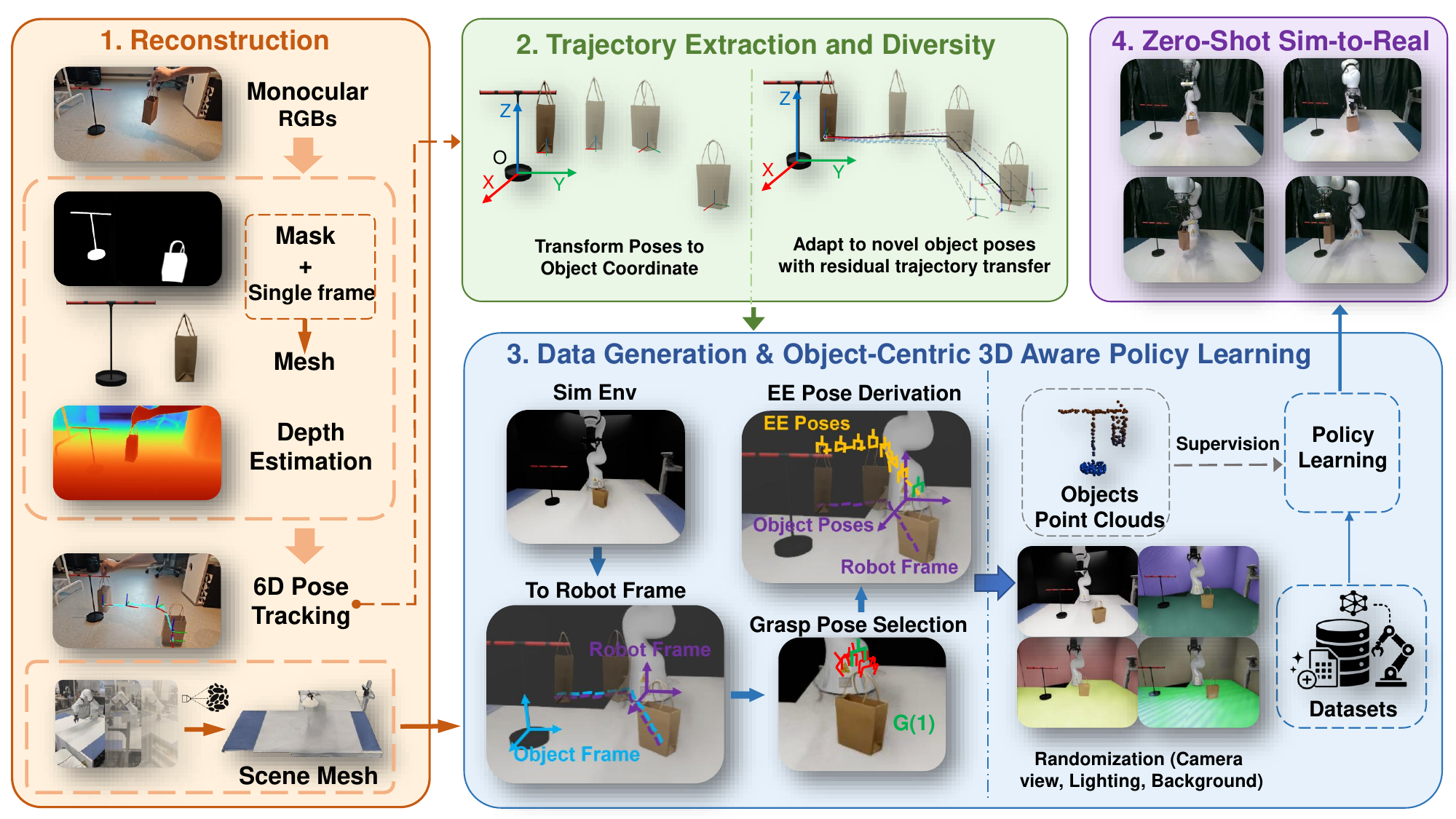}
    \caption{The pipeline of VidAct. VidAct reconstructs object meshes and trajectories from videos and transforms the trajectories into the object frame. Residual trajectory transfer efficiently augment the reconstructed trajectories to diverse object configurations. After selecting a grasp pose of gripper or Dexhand, VidAct derives the corresponding end-effector poses, rolls out the trajectories in simulation to generate training data, and trains an object-centric 3D-aware policy for zero-shot deployment.}
    \label{fig:pipeline}
    \vspace{-0.8em}
\end{figure*}

We propose VidAct, a video to robot system for learning robotic manipulation from arbitrary-view monocular RGB videos without action labels. As shown in Figure \ref{fig:pipeline}, VidAct consists of three stages: 1) \textbf{Video-to-Sim Reconstruction}: reconstructing object meshes and motion trajectories from monocular videos, and building photorealistic simulation scenes. 2) \textbf{Data Generation and Augmentation}: representing reconstructed trajectories with residual trajectory transfer for fast augmentation under novel object poses and generate robot-executable data demonstrations in simulation. 3) \textbf{Policy Learning and Sim-to-Real Evaluation}: training an image-conditioned policy to predict robot actions, with object-centric 3D point-cloud prediction as auxiliary supervision for transferring privileged simulation information, enabling zero-shot real-world deployment.

\subsection{Video-to-Sim Reconstruction}
\subsubsection{Objects Asset Extraction}
Given a monocular RGB video $\mathcal{V}=\{I_t\}_{t=1}^{T}$, we reconstruct the manipulated object mesh using SAM 3D~\cite{sam3d}. We first apply SAM2~\cite{sam2} to segment the manipulated object in each frame, obtaining masks $\{\Omega_t\}_{t=1}^{T}$. We then select the frame with the largest mask area and denote the corresponding image-mask pair as $(I^\star,\Omega^\star)$, which provides a complete visible object observation. This pair is fed into SAM 3D~\cite{sam3d} to reconstruct an initial object mesh $\tilde{\mathcal{O}}_m$ with an estimated scale.



\subsubsection{Trajectory Extraction}
To decouple the trajectory from the demonstrator's embodiment in video, we represent it with object poses rather than human or robot actions. To enable pose estimation with FoundationPose~\cite{foundationpose}, we first estimate metric depth maps $\{D_t\}_{t=1}^{T}$ using MoGe-3~\cite{MoGe-3}. We then align the reconstructed mesh $\tilde{\mathcal{O}}_m$ to the depth scale. For the selected frame, we use $D^\star$ and $\Omega^\star$ to obtain a point cloud $\mathcal{P}_o$ of the object. Taking $\mathcal{P}_o$ as the scale reference, we rescale $\tilde{\mathcal{O}}_m$ by matching the object size, yielding the near-metric object mesh $\mathcal{O}_m$. Given the RGB video $\mathcal{V}=\{I_t\}_{t=1}^{T}$, depth maps $\{D_t\}_{t=1}^{T}$, and object mesh $\mathcal{O}_m$, we use FoundationPose~\cite{foundationpose} to estimate the 6-DoF object pose sequence $\{\mathbf{P}_c^o(t)\}_{t=1}^{T}$ in the camera frame. To reduce noise, we sparsify the pose sequence into key poses and interpolate them into a smoother reference trajectory.

We assume a fixed camera within each demonstration video, while allowing arbitrary camera viewpoints across videos. To remove camera-viewpoint dependence, we transform each pose to the initial object frame: $\mathbf{P}^o(t)=\bigl(\mathbf{P}_c^o(1)\bigr)^{-1}\mathbf{P}_c^o(t)$, yielding the object-centric reference trajectory $\tau^o = \{\mathbf{P}^o(t)\}_{t=1}^{T}$. For two-object manipulation tasks (\textit{e.g.}, flower insertion or cup stacking), a single object-centric frame is insufficient because the task is defined by the spatial relation between two objects. We therefore use the static object as the reference frame and express the moving object's pose relative to it. Given the moving-object pose sequence $\{\mathbf{P}_c^m(t)\}_{t=1}^{T}$ and static-object pose $\mathbf{P}_c^s$, we compute $\mathbf{P}^{m|s}(t)=\bigl(\mathbf{P}_c^s\bigr)^{-1}\mathbf{P}_c^m(t)$, producing the object-to-object trajectory $\tau^{m|s} = \{\mathbf{P}^{m|s}(t)\}_{t=1}^{T}$.


\subsubsection{Environment Reconstruction}
While our framework does not necessarily require a reconstructed workspace, we optionally use 2D Gaussian Splatting~\cite{2dgs} to extract a mesh of the real scene, reducing visual discrepancies between simulation and real. After camera calibration, we import the scene and object meshes into Isaac Sim. To validate policy generalization, we apply domain randomization during data generation, including tabletop and background textures, lighting, camera poses, and object scales.

\subsection{Data Generation and Augmentation}
\begin{figure}[t]
    \centering
    \includegraphics[width=0.85\linewidth]{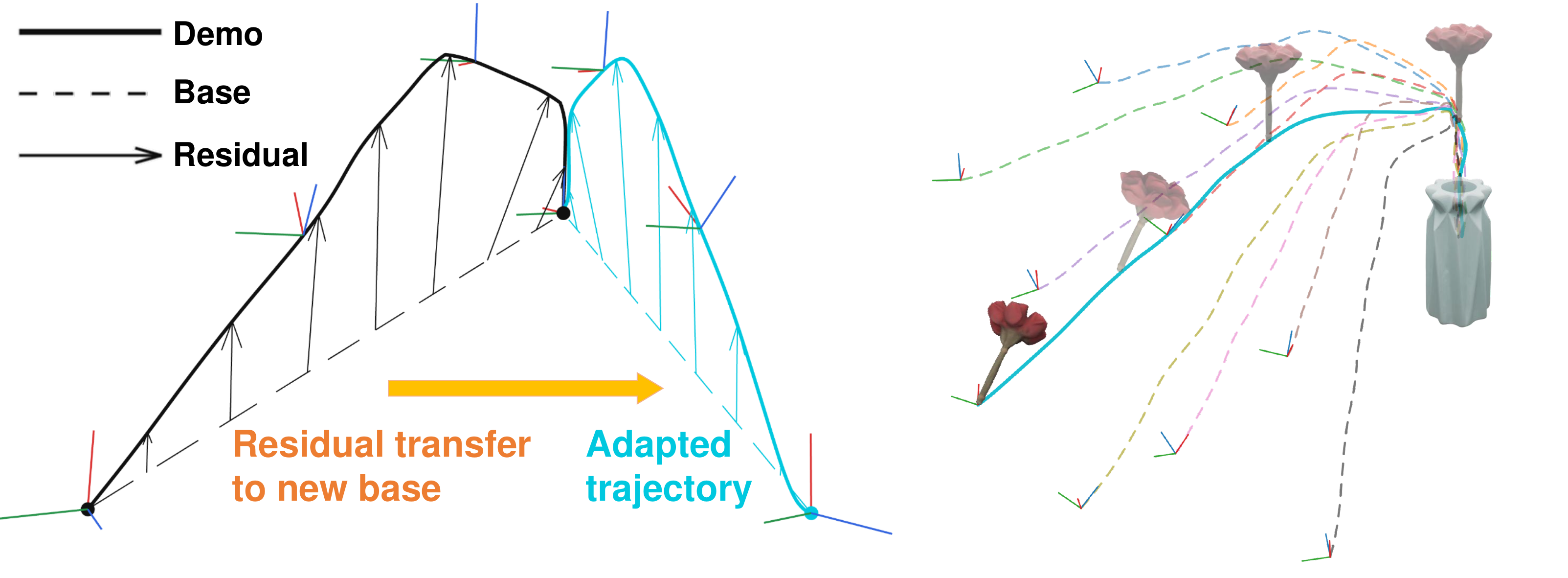}
    \caption{Residual trajectory transfer for trajectory augmentation. Left: illustration of residual trajectory transfer; right: sampled novel object poses.}
    \label{fig:dmp}
    \vspace{-15pt}
\end{figure}

\subsubsection{Residual Trajectory Transfer for Object Trajectory Diversity}
We use residual trajectory transfer to adapt
$\tau^{m|s}=\{\mathbf{P}^{m|s}(t)\}_{t=1}^{T}$ to a new initial pose. We decompose the demonstrated motion into a start-to-goal baseline and a corresponding residual. 

Specifically, we construct the baseline using linear position
interpolation and orientation SLERP. For a new initial pose, a new baseline is constructed, and the demonstrated residual is transferred onto it. The positional residual is aligned with the new approach direction and scaled according to the change in horizontal start-to-goal distance, while its vertical component is preserved. The orientation residual relative to the SLERP baseline is directly reused. This efficiently adapts the trajectory while retaining the demonstrated local motion pattern. Fig. \ref{fig:dmp} shows the illustration of residual trajectory transfer and visualization of some new trajectories generated by residual trajectory transfer.

\begin{figure}[t]
    \centering
    \includegraphics[width=0.85\linewidth]{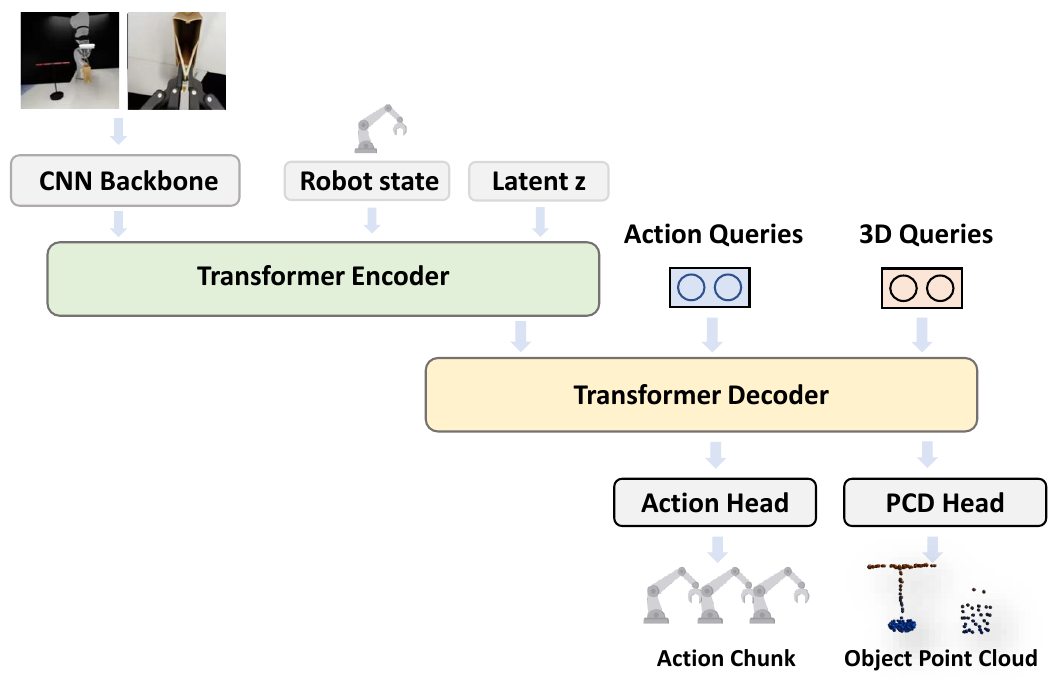}
    \caption{Model architecture of object-centric 3D predict policy. We add learnable query and auxiliary head to ACT \cite{ACT}.}
    \label{fig:model}
    \vspace{-15pt}
\end{figure}

\subsubsection{Object Trajectory to Robot Actions}
The reference trajectory $\tau^{m|s} = \{\mathbf{P}^{m|s}(t)\}_{t=1}^{T}$ encodes the moving object's pose relative to a static object frame. We first transform $\tau^{m|s}$ into the robot base frame as $\mathbf{P}_b^m(t) = \mathbf{P}_b^s \cdot \mathbf{P}^{m|s}(t)$, where $\mathbf{P}_b^s$ is the static object pose in the robot base frame.

To determine the end-effector trajectory, we generate candidate grasp poses for the moving object and select the optimal grasp pose $\mathbf{G}(1)$ at the first frame, which involves simulation testing and manual screening. We employ an automatic pipeline to generate candidate grasps from object geometry. For parallel grippers, grasp poses are sampled from opposing surface contacts under gripper width constraints. For dexterous hands, candidates are initialized from predefined grasp configurations and sampled over the object surface. All candidates are executed in simulation, and only successful grasp are retained for data generation.

Assuming the gripper/dexhand maintains a rigid grasp throughout execution, the relative transform from the moving object to the gripper is constant, $T_{obj \rightarrow ee} = \bigl(\mathbf{P}_b^m(1)\bigr)^{-1} \cdot \mathbf{G}(1)$. Applying this fixed transform to the full object pose sequence yields the desired end-effector trajectory $\xi^{ee} = \{\mathbf{P}_b^{ee}(t)\}_{t=1}^{T}$, where $\mathbf{P}_b^{ee}(t) = \mathbf{P}_b^m(t) \cdot T_{obj \rightarrow ee}$.
Each end-effector pose $\mathbf{P}_b^{ee}(t)$ is represented as a action $\mathbf{a}_t = [\mathbf{x}_t^{ee},\, g_t]$, where $\mathbf{x}_t^{ee} \in \mathbb{R}^6$ is the end-effector pose and $g_t$ is the gripper/dexhand command. The full action sequence $\mathcal{A} = \{\mathbf{a}_t\}_{t=1}^{T}$ is executed in simulation via inverse kinematics.

\subsubsection{Collecting Robot Training Data}
In simulation, we collect training tuples $\{(\mathbf{a}_t, \mathbf{c}_t \mid o_t)\}_{t=1}^{T}$, where $o_t$ is the RGB observation, and $\mathbf{a}_t$ and $\mathbf{c}_t$ are the action and object point cloud label. The point cloud $\mathbf{c}_t$ is obtained by uniformly sampling from the object meshes and expressed in the robot base frame, encoding the pose, and complete geometry of the objects.

\subsection{Policy Learning}
We first train an image-conditioned policy ACT ~\cite{ACT} for visuomotor control. Given the current RGB observation $o_t$, the policy predicts a future action chunk $\hat{\mathbf{A}}_t=\{\hat{\mathbf{a}}_{t+k}\}_{k=0}^{H-1}$, where $H$ is the action horizon. We further allow ACT to access privileged simulation information during training through object-centric auxiliary supervision. Specifically, we augment ACT with learnable tokens and an auxiliary head that predicts the current-frame point cloud $\hat{\mathbf{c}}_t$ of task-relevant objects. Since point clouds are unordered, we supervise $\hat{\mathbf{c}}_t$ with Chamfer loss, which measures the bidirectional nearest-neighbor distance between the predicted and target point sets. The final objective combines the standard ACT loss with the auxiliary point-cloud prediction loss:
\begin{equation}
\mathcal{L}
=
\mathcal{L}_{\mathrm{ACT}}
+
\lambda_{\mathrm{pc}}
\big[
d(\hat{\mathbf{c}}_t,\mathbf{c}_t)
+
d(\mathbf{c}_t,\hat{\mathbf{c}}_t)
\big],
\end{equation}
where
\begin{equation}
d(\mathbf{a},\mathbf{b})
=
\frac{1}{|\mathbf{a}|}
\sum_{\mathbf{p}\in\mathbf{a}}
\min_{\mathbf{q}\in\mathbf{b}}
\|\mathbf{p}-\mathbf{q}\|_2^2.
\end{equation}
where $\mathcal{L}_{\mathrm{ACT}}$ denotes the original ACT objective and $\lambda_{\mathrm{pc}}$ controls the weight of the point-cloud prediction loss. Figure \ref{fig:model} illustrates the detailed model architecture.


\section{Experiments}
\label{sec:experiments}
We conduct experiments to evaluate VidAct through the following questions:

(1). \textbf{Generality across video sources}: How well does VidAct handle videos from diverse sources? 

(2). \textbf{Data quality and zero-shot real-world deployment}: Can VidAct generate high-quality data for training visuomotor policies and enable zero-shot real-world deployment? 

(3). \textbf{Benefits of object-centric 3D prediction}: How does object-centric 3D prediction improve policy performance in simulation and real-world?

(4) \textbf{Advantages of residual trajectory transfer}: How does Residual trajectory transfer improve trajectory adaptation quality, data-generation reliability, and downstream policy performance?

(5). \textbf{Applicability to different robot embodiments}: Can VidAct be applied to different robot embodiments without modifying the core video-to-robot pipeline?

We conduct experiments on two embodiments: KUKA iiwa 14 7-DoF robot + Robotiq gripper for simulation and real world tasks, KUKA iiwa 14 7-DoF robot + Inspire RH56E2 five finger hand for simulation. The simulation environments are built in Isaac Sim. We use two Intel RealSense cameras at wrist and front with 480x640 resolution. Data generation, policy training and deployment are performed on a single RTX 5090. All the manipulation tasks are rigid-object manipulation.

\subsection{Generality across video sources}
\label{sec:video_source_generality}

We evaluate VidAct on videos from diverse sources, including human demonstrations (four self-collected videos and 20 human videos from PH2D \cite{PH2D}), 20 AI-generated videos using Wan2.2~\cite{wan}, and 22 Internet robot videos from BridgeData
V2~\cite{bridgedata_v2}, DreamDojo~\cite{dreamdojo} and UMI~\cite{umi}. 

\begin{figure}[t]
    \centering
    \includegraphics[width=\linewidth]{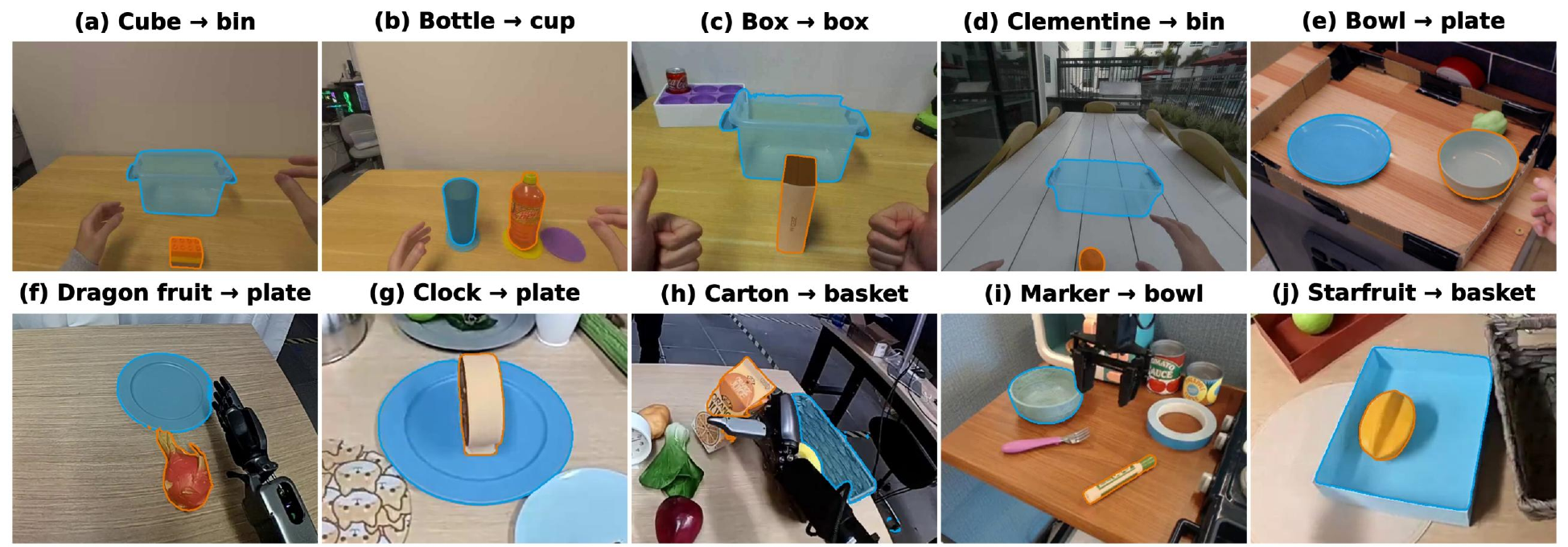}
    \vspace{-0.6em}
    \caption{Representative videos processed by VidAct.}
    \label{fig:video_source_examples}
\end{figure}

\begin{figure}[t]
    \centering
    \includegraphics[
        width=\linewidth,
        trim={0.1cm 0.4cm 0.1cm 0.4cm},
        clip
    ]{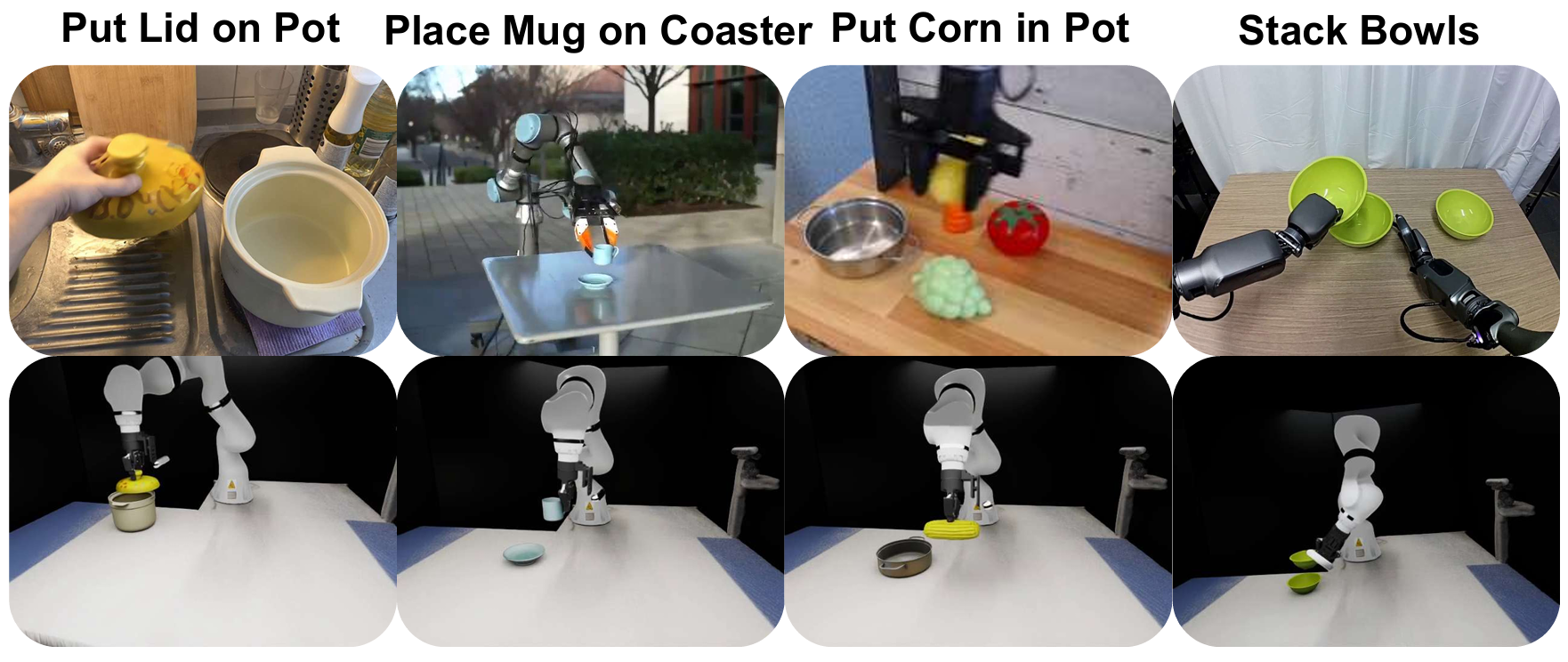}
    \vspace{-0.6em}
    \caption{Representative videos and the corresponding simulation tasks processed by VidAct. Top: Videos; bottom: Simulation. The video sources are generated from Wan2.2\cite{wan} and from UMI\cite{umi}, DreamDojo\cite{dreamdojo} and BridgeData V2\cite{bridgedata_v2}.}
    \label{fig:sim_examples}
    \vspace{-1.2em}
\end{figure}

Fig.~\ref{fig:video_source_examples} and
Fig.~\ref{fig:sim_examples} show representative input videos and reconstructed simulation tasks.

We report the video-to-simulation conversion rate for each video source, defining success as task completion by the robot in simulation. We further break down failures at three stages: object mask generation, mesh reconstruction, and pose tracking. Each stage is evaluated only on samples that pass the previous one. Table~\ref{tab:reliability} summarizes the results. 



\begin{table}[H]
    \centering
    \caption{Overall conversion rates across video sources and pipeline reliability across video sources.}
    \label{tab:reliability}
    \scriptsize
    \setlength{\tabcolsep}{4.5pt}
    \begin{tabular}{lc|ccc}
        \toprule
        \shortstack{Source Video Type} &
        Overall &
        \shortstack{Mask gen.} &
        \shortstack{Mesh recon.} &
        \shortstack{Pose tracking} \\
        \midrule
        Self-collected & 4/4 & 4/4 & 4/4   & 4/4 \\
        Internet Human & 14/20 & 19/20 & 16/19 & 14/16 \\
        Internet Robot & 15/22 & 20/22 & 18/20 & 15/18 \\
        Generated      & 9/20 & 17/20 & 14/17 & 9/14 \\
        \bottomrule
    \end{tabular}
    \vspace{-1.2em}
\end{table}


We evaluated the effect of scale alignment between the reconstructed mesh and estimated depth on pose tracking. Without scale alignment, the pose tracking success rate was nearly zero.


\begin{figure*}[!t]
    \centering
    \includegraphics[width=0.85\linewidth]{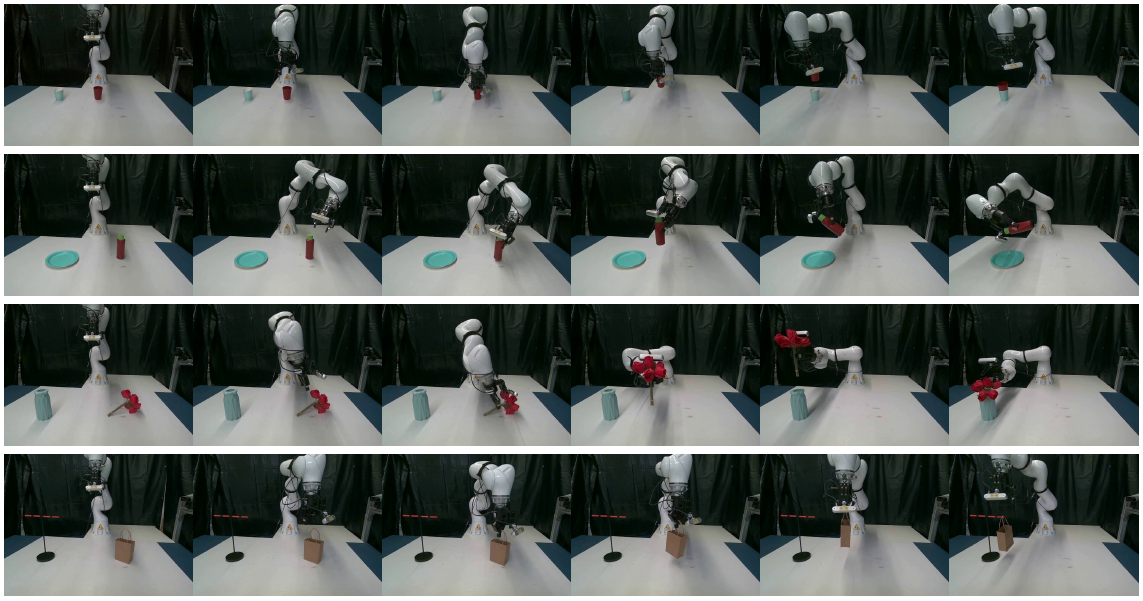}
    \caption{Zero-shot real-robot rollout visualizations for four tasks. From top to bottom, each row corresponds to Cup Stacking, Ketchup Pouring, Flower Insertion and Bag Hanging.}
    \label{fig:scaling_results}
    \vspace{-1.0em}
\end{figure*}

\begin{figure*}[t]
    \centering
    \includegraphics[width=\linewidth]{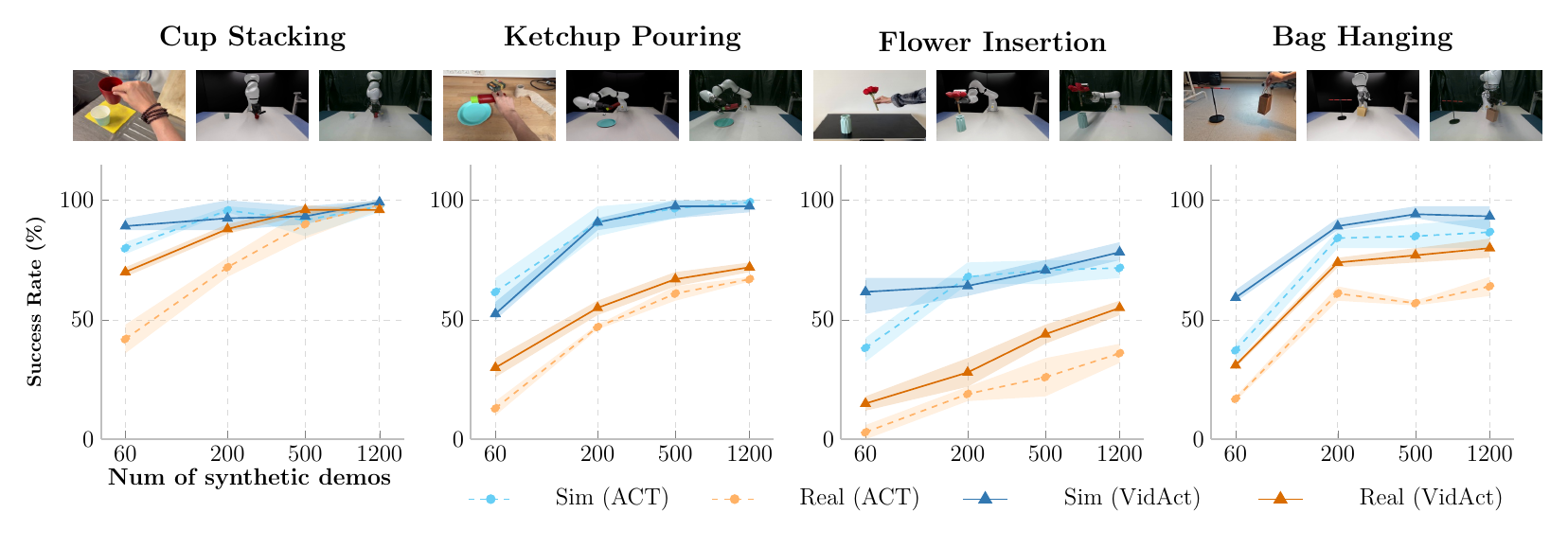}
    \vspace{-2.0em}
    \caption{Performance of standard visuomotor policy ACT~\cite{ACT} (2D) and our object-centric 3D prediction policy in simulation and the real world across varying sizes of training data. All policies are trained on simulated data without background randomization.}
    \label{fig:scaling_results}
    \vspace{-1.0em}
\end{figure*}

\subsection{Data quality and zero-shot real-world deployment}

To evaluate the quality of synthetic data generated through VidAct, we investigate the zero-shot transfer performance of standard visuomotor policy ACT~\cite{ACT}. Here, zero-shot deployment denotes that no real-robot demonstrations or real-world policy fine-tuning are used. Specifically, we benchmark the ACT across four manipulation tasks: Cup Stacking, Flower Insertion, Ketchup Pouring, and Paper Bag Hanging. For each task, we evaluate policies trained with three independent random seeds in simulation and two on the real robot, using 40 and 50 rollouts per seed, respectively. The object position randomization range is set to 30cm x 30cm. 

As shown in Fig.~\ref{fig:scaling_results}, ACT~\cite{ACT} achieves high simulation success rates on three of the four tasks using only 200 synthetic demonstrations. The policy can transfer to the real world, enabling zero-shot deployment, supporting the effectiveness of our generated training data.

\subsection{Benefits of object-centric 3D prediction}


To evaluate the benefits of our auxiliary object-centric 3D prediction, we compare the proposed object-centric 3D-aware policy against the ACT baseline. We augment ACT with learnable queries and a point-cloud prediction head, resulting in ACT\_3D. We first compare ACT and ACT\_3D under different training data scales, without background randomization, and evaluate their performance in both simulation and real-world settings. As shown in Fig.~\ref{fig:scaling_results}, although VidAct and the 2D baseline perform similarly in simulation, VidAct shows a consistent advantage in sim-to-real deployment, improving real-world success by \textbf{9.5}\% on average.






\subsubsection{Point cloud prediction accuracy}
\begin{figure}[t]
    \centering
    \includegraphics[width=\linewidth]{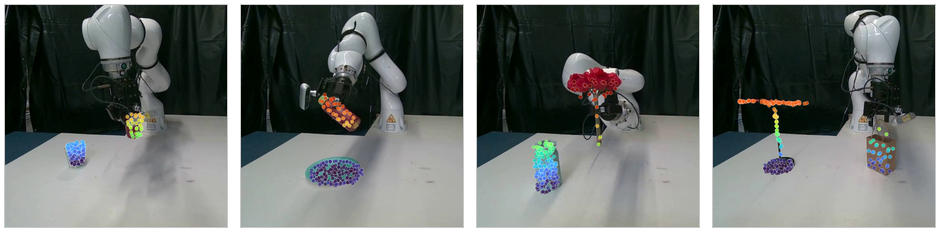}
    \caption{Visualization of real-world point cloud predictions projected onto the images.}
    \label{fig:real-pcd}
\end{figure}

We evaluated squared-$L_2$ Chamfer Distance between predicted and ground-truth point clouds over 100 episodes per task (400 total) in simulation, obtaining consistent values of $\mathbf{1x10^{-4}\text{--}4x10^{-4}}$ $\mathrm{m}^2$. In real-world experiments, ground-truth object point clouds are unavailable. Fig. \ref{fig:real-pcd} shows the visualizations of predicted object point cloud that are projected to image based on camera parameters.

\begin{figure}[t]
    \centering
    \includegraphics[
        width=\linewidth,
        trim={0.1cm 0.4cm 0.1cm 0.4cm},
        clip
    ]{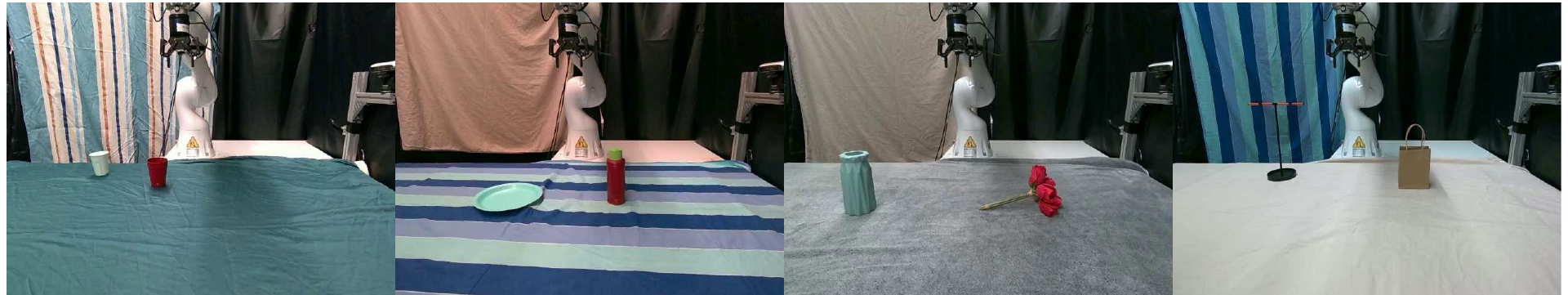}
    \caption{Generalization setup with diverse backgrounds and varied table textures and colors in real world.}
    \label{fig:real_generalization}
    \vspace{-1.0em}
\end{figure}

\subsubsection{Generalization to Unseen Scenes}
We collect 500 simulated trajectories per task with randomized backgrounds and textures, and train ACT, ACT\_Pose, and ACT\_3D on the same data. ACT\_Pose uses object-pose auxiliary supervision. We evaluate each method in unseen scenes with 200 simulation trials and 30 real-world trials per task. Figure~\ref{fig:real_generalization} shows representative generalization setups for evaluation.
\vspace{-1.0em}

\begin{table}[H]
    \centering
    \caption{Generalization comparison of ACT, ACT\_pose and ACT\_3D. The best result is \textbf{bold}.}
    \label{tab:generalization}
    \scriptsize
    \setlength{\tabcolsep}{2pt}
    \begin{tabularx}{\columnwidth}{
        @{}l|
        *{3}{>{\centering\arraybackslash}X}|
        *{3}{>{\centering\arraybackslash}X}
        @{}
    }
        \toprule
        & \multicolumn{3}{c|}{Simulation}
        & \multicolumn{3}{c}{Real World} \\
        \cmidrule(lr){2-4}
        \cmidrule(lr){5-7}
        Task & ACT & ACT\_Pose & ACT\_3D
             & ACT & ACT\_Pose & ACT\_3D \\
        \midrule
        Cup Stacking      & 88\%  & 89.5\% & \textbf{98\%} & 17/30 & 18/30 & \textbf{21/30} \\
        Ketchup Pouring   & 77.5\% & 79\% & \textbf{93.5\%} & 10/30 & 11/30 & \textbf{15/30} \\
        Flower Insertion  & 60\% & 58.5\% & \textbf{71\%} & 6/30 & 6/30 & \textbf{10/30} \\
        Bag Hanging       & 63.5\% & 65\% & \textbf{76.5\%} & 8/30 & 9/30 & \textbf{12/30} \\
        \bottomrule
    \end{tabularx}
    \vspace{-1.0em}
\end{table}

As shown in Table~\ref{tab:generalization}, ACT\_3D improves average success rates over ACT by \textbf{12.50} and \textbf{14.17} percentage points in simulation and the real world, respectively, and over ACT\_Pose by \textbf{11.75} and \textbf{11.67} points. These results support the benefits of complete-object 3D supervision over pose-only supervision for generalization to unseen scenes.

\subsection{Advantages of Residual Trajectory Transfer}

We compare Residual Trajectory Transfer with (1) trajectory interpolation from
Real2Render2Real~\cite{real2render2real} and (2) segmented skill replay
with motion planning from DemoGen~\cite{demogen}.
For each task, all methods generate trajectories within the same randomization region until 500 successful trajectories are collected. We report the trajectory-generation
success rate over all attempts and evaluate successful trajectories
using demo-shape RMSE(Demo-Shape RMSE measures the RMSE between path-length-normalized pairwise-distance matrices, invariant to translation, rotation, and uniform scaling.) and jerk RMS. To evaluate downstream policy performance, we further train ACT on 500 generated trajectories for two tasks and report simulation success rates.

\begin{table}[t]
\centering
\scriptsize
\setlength{\tabcolsep}{2pt}
\renewcommand{\arraystretch}{1.08}
\caption{Comparison of residual trajectory transfer with
trajectory augmentation baselines. The best result in each
row is \textbf{bold}.}
\label{tab:traj_aug}

\begin{tabular*}{\columnwidth}{@{\extracolsep{\fill}}lccc@{}}
\toprule
Task & Interpolation & Replay &
\shortstack{Residual Traj. Transfer (Ours)} \\
\midrule

\multicolumn{4}{@{}l@{}}{
    \textit{Trajectory Generation Success Rate}} \\
\addlinespace[2pt]
Cup Stacking
    & 69.1\% & 70.6\% & \textbf{96.4\%} \\
Ketchup Pouring
    & 83.6\% & 39.7\% & \textbf{95.8\%} \\
Flower Insertion
    & 24.8\% & 33.2\% & \textbf{85.4\%} \\
Bag Hanging
    & 17.0\% & 81.1\% & \textbf{98.2\%} \\
\addlinespace[2pt]
Average
    & 48.63\% & 56.15\% & \textbf{93.95\%} \\
\midrule

\multicolumn{4}{@{}l@{}}{
    \textit{Trajectory Quality}} \\
\multicolumn{4}{@{}l@{}}{
    \textit{Four-task average; successful trajectories only}} \\
\addlinespace[2pt]
Demo-Shape RMSE $\downarrow$
    & 0.148 & 0.105 & \textbf{0.068} \\
Jerk RMS ($\times 10^{3}$) $\downarrow$
    & 4.79 & 4.05 & \textbf{3.60} \\
\midrule

\multicolumn{4}{@{}l@{}}{
    \textit{ACT Simulation Success Rate}} \\
\addlinespace[2pt]
Cup Stacking
    & 58\% & 83\% & \textbf{92\%} \\
Bag Hanging
    & 24\% & 75\% & \textbf{87\%} \\
\addlinespace[2pt]
Average
    & 41\% & 79\% & \textbf{89.5\%} \\
\bottomrule
\end{tabular*}
\end{table}

Table~\ref{tab:traj_aug} shows that our method achieves the highest data generation success (93.95\%), lowest demo-shape RMSE (0.068), and highest average ACT success (89.5\%), supporting its benefits for data generation and downstream policy learning.



\subsection{Applicability to Different Robot Embodiments}
\begin{figure}[t]
    \centering
    \includegraphics[
        width=\linewidth,
        trim={0.1cm 0.4cm 0.1cm 0.4cm},
        clip
    ]{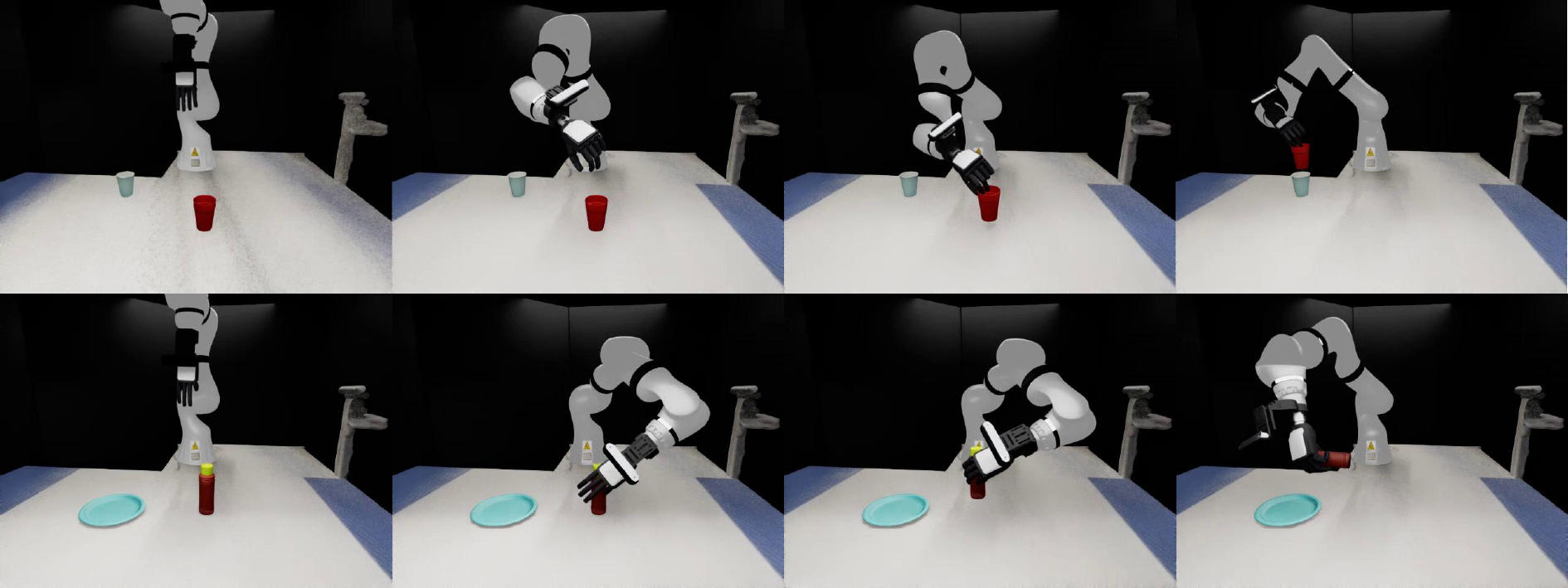}
    \caption{Dexterous hand rollouts in simulation.}
    \label{fig:dexterous}
    \vspace{-3.5mm}
\end{figure}

We test VidAct with a dexterous hand on Cup Stacking and Ketchup Pouring in simulation. The core pipeline remains unchanged; only the robot hardware configuration and grasp poses are adapted to the dexterous hand. As shown in Fig.~\ref{fig:dexterous}, VidAct generates executable dexterous-hand rollouts. Using 200 generated trajectories per task, ACT\_3D achieves success rates of \textbf{67.5}\% and \textbf{77.5}\% over 200 evaluation trials on Cup Stacking and Ketchup Pouring, respectively.

\section{CONCLUSION}
\label{sec:conclusion}
We propose VidAct, a video-to-robot system for learning object-centric 3D-aware manipulation policies from one monocular video per task, enabling zero-shot real-world deployment. By treating object motion as the transferable trajectory and transforming it into the object frame, VidAct removes dependence on video viewpoint and demonstrator embodiment. VidAct further adapts reference trajectories with residual transfer, enabling fast trajectory augmentation under novel object poses while preserving the trajectory shape. During policy learning, VidAct exploits privileged simulation information through an object-centric 3D prediction head, improving policy success rate and generalization performance without requiring 3D inputs at deployment. Experiments across diverse videos, simulated tasks, and four real-world tasks validate the VidAct.

\textbf{Limitation and future work} The current scope of VidAct is limited to rigid-object manipulation. Although VidAct supports diverse video sources, the video reconstruction pipeline is not always reliable. If higher video usability is desired, manual intervention will be needed for target-object selection and pose-tracking verification. Future work could make the pipeline more reliable and automated, e.g., by integrating an agentic framework for automatic inspection and failure handling. 

\balance
\bibliographystyle{IEEEtran}
\bibliography{chapter/reference}

@article{kareer2025emergence,
  title   = {Emergence of Human to Robot Transfer in Vision-Language-Action Models},
  author  = {Simar Kareer and Karl Pertsch and James Darpinian and Judy Hoffman and Danfei Xu and Sergey Levine and Chelsea Finn and Suraj Nair},
  journal = {arXiv preprint arXiv:2512.22414},
  year    = {2025},
  doi     = {10.48550/arXiv.2512.22414},
}

@article{EgoScale,
  title   = {EgoScale: Scaling Dexterous Manipulation with Diverse Egocentric Human Data},
  author  = {Ruijie Zheng and Dantong Niu and Yuqi Xie and Jing Wang and Mengda Xu and Yunfan Jiang and Fernando Castañeda and Fengyuan Hu and You Liang Tan and Letian Fu and Trevor Darrell and Furong Huang and Yuke Zhu and Danfei Xu and Linxi Fan},
  journal = {arXiv preprint arXiv:2602.16710},
  year    = {2026},
  url     = {https://arxiv.org/abs/2602.16710}
}

@inproceedings{egodex,
  title     = {EgoDex: Learning Dexterous Manipulation from Large-Scale Egocentric Video},
  author    = {Hoque Ryan and Huang Peide and Yoon David J. and Sivapurapu Mouli and Zhang Jian},
  booktitle = {International Conference on Learning Representations (ICLR)},
  year      = {2026}
}

@inproceedings{real2render2real,
  title     = {Real2Render2Real: Scaling Robot Data Without Dynamics Simulation or Robot Hardware},
  author    = {Justin Yu and Letian Fu and Huang Huang and Karim El-Refai and Rares Andrei Ambrus and Richard Cheng and Muhammad Zubair Irshad and Ken Goldberg},
  booktitle = {Conference on Robot Learning (CoRL)},
  year      = {2025},
}

@inproceedings{xsim,
  title     = {X-Sim: Cross-Embodiment Learning via Real-to-Sim-to-Real},
  author    = {Prithwish Dan and Kushal Kedia and Angela Chao and Edward Weiyi Duan and Maximus Adrian Pace and Wei-Chiu Ma and Sanjiban Choudhury},
  booktitle = {Conference on Robot Learning (CoRL)},
  year      = {2025},
}

@inproceedings{rialto,
  title     = {Reconciling Reality through Simulation: A Real-To-Sim-to-Real Approach for Robust Manipulation},
  author    = {Torne Villasevil, Marcel and Simeonov, Anthony and Li, Zechu and Chan, April and Chen, Tao and Gupta, Abhishek and Agrawal, Pulkit},
  booktitle = {Robotics: Science and Systems (RSS)},
  year      = {2024},
  doi       = {10.15607/RSS.2024.XX.015}
}

@inproceedings{mimicgen,
  title     = {MimicGen: A Data Generation System for Scalable Robot Learning using Human Demonstrations},
  author    = {Mandlekar Ajay and Nasiriany Soroush and Wen Bowen and Akinola Iretiayo and Narang Yashraj and Fan Linxi and Zhu Yuke and Fox Dieter},
  booktitle = {Conference on Robot Learning (CoRL)},
  year      = {2023}
}

@inproceedings{demogen,
  title     = {DemoGen: Synthetic Demonstration Generation for Data-Efficient Visuomotor Policy Learning},
  author    = {Xue Zhengrong and Deng Shuying and Chen Zhenyang and Wang Yixuan and Yuan Zhecheng and Xu Huazhe},
  booktitle = {Robotics: Science and Systems (RSS)},
  year      = {2025},
  doi       = {10.15607/RSS.2025.XXI.157}
}

@inproceedings{ACT,
	title={Learning Fine-Grained Bimanual Manipulation with Low-Cost Hardware},
	author={Tony Z. Zhao and Vikash Kumar and Sergey Levine and Chelsea Finn},
	booktitle={RSS},
	year={2023}
}

@inproceedings{Diffusion-Policy,
	title={Diffusion Policy: Visuomotor Policy Learning via Action Diffusion},
	author={Cheng Chi and Zhenjia Xu and Siyuan Feng and Eric Cousineau and Yilun Du and Benjamin Burchfiel and Russ Tedrake and Shuran Song},
	booktitle={RSS},
	year={2023}
}

@inproceedings{pi0.5,
 author = {Physical Intelligence},
 title = {{$\pi$}0.5: a Vision-Language-Action Model with Open-World Generalization},
 booktitle = CoRL,
 note = {arXiv:2504.16054},
 year = 2025
}

@article{deximit,
  title   = {DexImit: Learning Bimanual Dexterous Manipulation from Monocular Human Videos},
  author  = {Mu Juncheng and Yang Sizhe and Bao Yiming and Bae Hojin and Wei Tianming and Xu Linning and Li Boyi and Xu Huazhe and Pang Jiangmiao},
  journal = {arXiv preprint arXiv:2602.10105},
  year    = {2026},
  url     = {https://arxiv.org/abs/2602.10105}
}

@inproceedings{3D-Diffusion-Policy,
	title={3D Diffusion Policy: Generalizable Visuomotor Policy Learning via Simple 3D Representations},
	author={Yanjie Ze and Gu Zhang and Kangning Zhang and Chenyuan Hu and Muhan Wang and Huazhe Xu},
	booktitle={RSS},
	year={2024}
}

@inproceedings{o3dp,
	title={Language-Guided Object-Centric Diffusion Policy for Generalizable and Collision-Aware Robotic Manipulation},
	author={Hang Li and Qian Feng and Zhi Zheng and Jianxiang Feng and Zhaopeng Chen and Alois Knoll},
	booktitle={ICRA},
	year={2025}
}

@article{dexman,
  title   = {DexMan: Learning Bimanual Dexterous Manipulation from Human and Generated Videos},
  author  = {Jhen Hsieh and Kuan-Hsun Tu and Kuo-Han Hung and Tsung-Wei Ke},
  journal = {arXiv preprint arXiv:2510.08475},
  year    = {2025},
}

@inproceedings{rigvid,
  title     = {{RIGVid}: Robotic Manipulation by Imitating Generated Videos Without Physical Demonstrations},
  author    = {Shivansh Patel and Shraddhaa Mohan and Hanlin Mai and Unnat Jain and Svetlana Lazebnik and Yunzhu Li},
  booktitle = {International Conference on Learning Representations (ICLR)},
  year      = {2026}
}

@inproceedings{viola,
  title     = {{VIOLA}: Imitation Learning for Vision-Based Manipulation with Object Proposal Priors},
  author    = {Yifeng Zhu and Abhishek Joshi and Peter Stone and Yuke Zhu},
  booktitle = {Conference on Robot Learning (CoRL)},
  year      = {2023}
}

@inproceedings{vima,
  title     = {{VIMA}: General Robot Manipulation with Multimodal Prompts},
  author    = {Yunfan Jiang and Agrim Gupta and Zichen Zhang and Guanzhi Wang and Yongqiang Dou and Yanjun Chen and Li Fei-Fei and Anima Anandkumar and Yuke Zhu and Linxi Fan},
  booktitle = {International Conference on Machine Learning (ICML)},
  year      = {2023}
}

@inproceedings{moo,
  title     = {Open-World Object Manipulation using Pre-Trained Vision-Language Models},
  author    = {Austin Stone and Ted Xiao and Yao Lu and Keerthana Gopalakrishnan and Kuang-Huei Lee and Quan Vuong and Paul Wohlhart and Sean Kirmani and Brianna Zitkovich and Fei Xia and Chelsea Finn and Karol Hausman},
  booktitle = {Conference on Robot Learning (CoRL)},
  year      = {2023}
}

@inproceedings{multitask,
  title     = {Multi-Task Domain Adaptation for Deep Learning of Instance Grasping from Simulation},
  author    = {Kuan Fang and Yunfei Bai and Stefan Hinterstoisser and Silvio Savarese and Mrinal Kalakrishnan},
  booktitle = {IEEE International Conference on Robotics and Automation (ICRA)},
  year      = {2018},
}

@inproceedings{groot,
  title     = {Learning Generalizable Manipulation Policies with Object-Centric 3D Representations},
  author    = {Yifeng Zhu and Zhenyu Jiang and Peter Stone and Yuke Zhu},
  booktitle = {Conference on Robot Learning (CoRL)},
  year      = {2023}
}

@inproceedings{ecot,
  title     = {Robotic Control via Embodied Chain-of-Thought Reasoning},
  author    = {Micha{\l} Zawalski and William Chen and Karl Pertsch and Oier Mees and Chelsea Finn and Sergey Levine},
  booktitle = {Conference on Robot Learning (CoRL)},
  year      = {2025}
}

@article{flowvla,
  title   = {{FlowVLA}: Thinking in Motion with a Visual Chain of Thought},
  author  = {Zhide Zhong and Haodong Yan and Junfeng Li and Xiangchen Liu and Xin Gong and Wenxuan Song and Jiayi Chen and Haoang Li},
  journal = {arXiv preprint arXiv:2508.18269},
  year    = {2025},
}

@article{lamp,
  title   = {{LaMP}: Learning Vision-Language-Action Policies with 3D Scene Flow as Latent Motion Prior},
  author  = {Xinkai Wang and Chenyi Wang and Yifu Xu and Mingzhe Ye and Fu-Cheng Zhang and Jialin Tian and Xinyu Zhan and Lifeng Zhu and Cewu Lu and Lixin Yang},
  journal = {arXiv preprint arXiv:2603.25399},
  year    = {2026},
}

@article{qdepthvla,
  title   = {{QDepth-VLA}: Quantized Depth Prediction as Auxiliary Supervision for Vision-Language-Action Models},
  author  = {Yixuan Li and Yuhui Chen and Mingcai Zhou and Haoran Li and Zhengtao Zhang and Dongbin Zhao},
  journal = {arXiv preprint arXiv:2510.14836},
  year    = {2025},
}

@inproceedings{dreamvla,
  title     = {{DreamVLA}: A Vision-Language-Action Model Dreamed with Comprehensive World Knowledge},
  author    = {Wenyao Zhang and Hongsi Liu and Zekun Qi and Yunnan Wang and Xinqiang Yu and Jiazhao Zhang and Runpei Dong and Jiawei He and Fan Lu and He Wang and Zhizheng Zhang and Li Yi and Wenjun Zeng and Xin Jin},
  booktitle = {Advances in Neural Information Processing Systems (NeurIPS)},
  year      = {2025}
}

@article{robointer,
  title   = {{RoboInter}: A Holistic Intermediate Representation Suite Towards Robotic Manipulation},
  author  = {Hao Li and Ziqin Wang and Zi-han Ding and Shuai Yang and Yilun Chen and Yang Tian and Xiaolin Hu and Tai Wang and Dahua Lin and Feng Zhao and Si Liu and Jiangmiao Pang},
  journal = {arXiv preprint arXiv:2602.09973},
  year    = {2026},
}

@inproceedings{dexmimicgen,
  title     = {{DexMimicGen}: Automated Data Generation for Bimanual Dexterous Manipulation via Imitation Learning},
  author    = {Zhenyu Jiang and Yuqi Xie and Kevin Lin and Zhenjia Xu and Weikang Wan and Ajay Mandlekar and Linxi Fan and Yuke Zhu},
  booktitle = {IEEE International Conference on Robotics and Automation (ICRA)},
  year      = {2025},
}

@inproceedings{sam2,
  title     = {{SAM} 2: Segment Anything in Images and Videos},
  author    = {Nikhila Ravi and Valentin Gabeur and Yuan-Ting Hu and Ronghang Hu and Chaitanya Ryali and Tengyu Ma and Haitham Khedr and Roman R{\"a}dle and Chloe Rolland and Laura Gustafson and Eric Mintun and Junting Pan and Kalyan Vasudev Alwala and Nicolas Carion and Chao-Yuan Wu and Ross Girshick and Piotr Dollar and Christoph Feichtenhofer},
  booktitle = {International Conference on Learning Representations (ICLR)},
  year      = {2025}
}

@article{sam3d,
  title   = {{SAM} 3D: 3Dfy Anything in Images},
  author  = {{SAM 3D Team} and Xingyu Chen and Fu-Jen Chu and Pierre Gleize and Kevin J Liang and Alexander Sax and Hao Tang and Weiyao Wang and Michelle Guo and Thibaut Hardin and Xiang Li and Aohan Lin and Jiawei Liu and Ziqi Ma and Anushka Sagar and Bowen Song and Xiaodong Wang and Jianing Yang and Bowen Zhang and Piotr Doll{\'a}r and Georgia Gkioxari and Matt Feiszli and Jitendra Malik},
  journal = {arXiv preprint arXiv:2511.16624},
  year    = {2025},
}

@inproceedings{foundationpose,
  title     = {{FoundationPose}: Unified 6D Pose Estimation and Tracking of Novel Objects},
  author    = {Bowen Wen and Wei Yang and Jan Kautz and Stan Birchfield},
  booktitle = {IEEE/CVF Conference on Computer Vision and Pattern Recognition (CVPR)},
  year      = {2024}
}

@inproceedings{2dgs,
  title     = {2D Gaussian Splatting for Geometrically Accurate Radiance Fields},
  author    = {Binbin Huang and Zehao Yu and Anpei Chen and Andreas Geiger and Shenghua Gao},
  booktitle = {ACM SIGGRAPH Conference Papers},
  year      = {2024},
  doi       = {10.1145/3641519.3657428}
}

@misc{wan,
  title={Wan: Open and Advanced Large-Scale Video Generative Models},
  author={{Wan Team}},
  year={2025},
  eprint={2503.20314},
  archivePrefix={arXiv}
}

@inproceedings{umi,
  title={Universal Manipulation Interface: In-the-Wild Robot Teaching Without In-the-Wild Robots},
  author={Chi, Cheng and Xu, Zhenjia and Pan, Chuer and Cousineau, Eric and Burchfiel, Benjamin and Feng, Siyuan and Tedrake, Russ and Song, Shuran},
  booktitle={Robotics: Science and Systems},
  year={2024}
}

@inproceedings{bridgedata_v2,
  title={BridgeData V2: A Dataset for Robot Learning at Scale},
  author={Walke, Homer and Black, Kevin and Lee, Abraham and Kim, Moo Jin and Du, Max and Zheng, Chongyi and Zhao, Tony and Hansen-Estruch, Philippe and Vuong, Quan and He, Andre and Myers, Vivek and Fang, Kuan and Finn, Chelsea and Levine, Sergey},
  booktitle={Conference on Robot Learning},
  year={2023}
}

@misc{dreamdojo,
  title={DreamDojo: A Generalist Robot World Model from Large-Scale Human Videos},
  author={Gao, Shenyuan and Liang, William and Zheng, Kaiyuan and Malik, Ayaan and Ye, Seonghyeon and Yu, Sihyun and Tseng, Wei-Cheng and Dong, Yuzhu and Mo, Kaichun and Lin, Chen-Hsuan and Ma, Qianli and Nah, Seungjun and Magne, Loic and Xiang, Jiannan and Xie, Yuqi and Zheng, Ruijie and Niu, Dantong and Tan, You Liang and Zentner, K. R. and Kurian, George and Indupuru, Suneel and Jannaty, Pooya and Gu, Jinwei and Zhang, Jun and Malik, Jitendra and Abbeel, Pieter and Liu, Ming-Yu and Zhu, Yuke and Jang, Joel and Fan, Linxi},
  year={2026},
  eprint={2602.06949},
  archivePrefix={arXiv}
}

@misc{MoGe-3,
  title={MoGe-3: Fine-Detail Monocular Geometry Estimation with Self-Guided Sparse Volumetric Refinement},
  author={Lingyu Kong and Ruicheng Li and Ruicheng Wang and Sicheng Xu and Chengtang Yao and Jianfeng Xiang and Jiaolong Yang},
  year={2026},
  eprint={2607.17967},
  archivePrefix={arXiv}
}

@misc{PH2D,
  title={Humanoid Policy {$\sim$} Human Policy},
  author={Ri-Zhao Qiu and Shiqi Yang and Xuxin Cheng and Chaitanya Chawla and Jialong Li and Tairan He and Ge Yan and David J. Yoon and Ryan Hoque and Lars Paulsen and Ge Yang and Jian Zhang and Sha Yi and Guanya Shi and Xiaolong Wang},
  year={2025},
  eprint={2503.13441},
  archivePrefix={arXiv}
}

@misc{ReMem-VLA,
  title={ReMem-VLA: Empowering Vision-Language-Action Model with Memory via Dual-Level Recurrent Queries},
  author={Hang Li and Fengyi Shen and Dong Chen and Liudi Yang and Xudong Wang and Jinkui Shi and Zhenshan Bing and Ziyuan Liu and Alois Knoll},
  year={2026},
  eprint={2603.12942},
  archivePrefix={arXiv}
}

\end{document}